\documentclass[10pt,journal,compsoc]{IEEEtran}

\ifCLASSOPTIONcompsoc
  \usepackage[nocompress]{cite}
\else
  \usepackage{cite}
\fi

\usepackage{newfloat}
\DeclareFloatingEnvironment[fileext=lop]{Algorithm}

\usepackage{color}
\usepackage{epsfig}
\usepackage{graphicx}
\usepackage{amsmath}
\usepackage{amssymb}
\usepackage{adjustbox}
\usepackage{multirow}
\usepackage{verbatim}
\usepackage{tablefootnote}
\usepackage{framed}
\usepackage{tikz}
 \def\checkmark{\tikz\fill[scale=0.4](0,.35) -- (.25,0) -- (1,.7) -- (.25,.15) -- cycle;}

\newcommand{\app}{\raise.17ex\hbox{$\scriptstyle\sim$}}

\ifCLASSINFOpdf
\else
\fi

\begin{document}

\title{Learning Two-Branch Neural Networks for Image-Text Matching Tasks}
\author{Liwei Wang, Yin Li, Jing Huang, Svetlana Lazebnik
\IEEEcompsocitemizethanks{\IEEEcompsocthanksitem Liwei Wang, Jing Huang and Svetlana Lazebnik are with the Computer Science Department, University of Illinois at Urbana-Champaign, Urbana, IL 61801. E-mail: lwang97@illinois.edu, jhuang81@illinois.edu and slazebni@illinois.edu\protect \\
\IEEEcompsocthanksitem Yin Li is with the School of Interactive Computing, Georgia Institute of Technology, Atlanta, GA, 30332.E-mail: yli440@gatech.edu \protect\\
}}


\IEEEtitleabstractindextext{%
\begin{abstract}
Image-language matching tasks have recently attracted a lot of attention in the computer vision field. These tasks include image-sentence matching, i.e., given an image query, retrieving relevant sentences and vice versa, and region-phrase matching or visual grounding, i.e., matching a phrase to relevant regions. This paper investigates two-branch neural networks for learning the similarity between these two data modalities. We propose two network structures that produce different output representations. The first one, referred to as an {\em embedding network}, learns an explicit shared latent embedding space with a maximum-margin ranking loss and novel neighborhood constraints. Compared to standard triplet sampling, we perform improved neighborhood sampling that takes neighborhood information into consideration while constructing mini-batches. The second network structure, referred to as a {\em similarity network}, fuses the two branches via element-wise product and is trained with regression loss to directly predict a similarity score. Extensive experiments show that our networks achieve high accuracies for phrase localization on the Flickr30K Entities dataset and for bi-directional image-sentence retrieval on Flickr30K and MSCOCO datasets.
\end{abstract}

\begin{IEEEkeywords} Deep Learning, Cross-Modal Retrieval, Image-Sentence Retrieval, Phrase Localization, Visual Grounding
\end{IEEEkeywords}}

\maketitle
\IEEEdisplaynontitleabstractindextext
\IEEEpeerreviewmaketitle
\ifCLASSOPTIONcompsoc
\IEEEraisesectionheading{\section{Introduction}\label{sec:introduction}}
\else
\section{Introduction}
\label{sec:introduction}
\fi

\IEEEPARstart{C}{omputer} vision is moving from predicting discrete, categorical labels to generating rich descriptions of visual data, in particular, in the form of natural language. We are witnessing a surge of interest in tasks that involve cross-modal learning from image and text data, widely viewed as the ``next frontier'' of scene understanding. For example, in bi-directional image-sentence search~\cite{karpathy2014deep,klein2014fisher,gong2014improving} one aims to retrieve the corresponding images given a sentence query, and vice versa. Image captioning~\cite{karpathy2015deep,johnson2015densecap,vinyals2015show} is the task of generating a natural language description of an input image. Motivated by the notion of creating a visual Turing test, Visual Question Answering (VQA)~\cite{antol2015vqa,yu2015visual,jabri2016revisiting} aims at answering freeform questions about image content. Visual grounding tasks like referring expression understanding~\cite{kazemzadeh2014referitgame,yu2016modeling} and phrase localization~\cite{plummer2015flickr30k} find image regions indicated by questions, phrases, or sentences. To support these tasks, a number of large-scale datasets and benchmarks have recently been proposed, including MSCOCO~\cite{chen2015microsoft} and Flickr30K~\cite{young2014image} datasets for image captioning, Flickr30K Entities~\cite{plummer2015flickr30k} for phrase localization, the Visual Genome dataset~\cite{krishnavisualgenome} for localized textual description of images, and the VQA dataset~\cite{antol2015vqa} for question answering.

\begin{figure*}[t]
\hspace{0cm}  
\includegraphics[width=1.0\linewidth]
{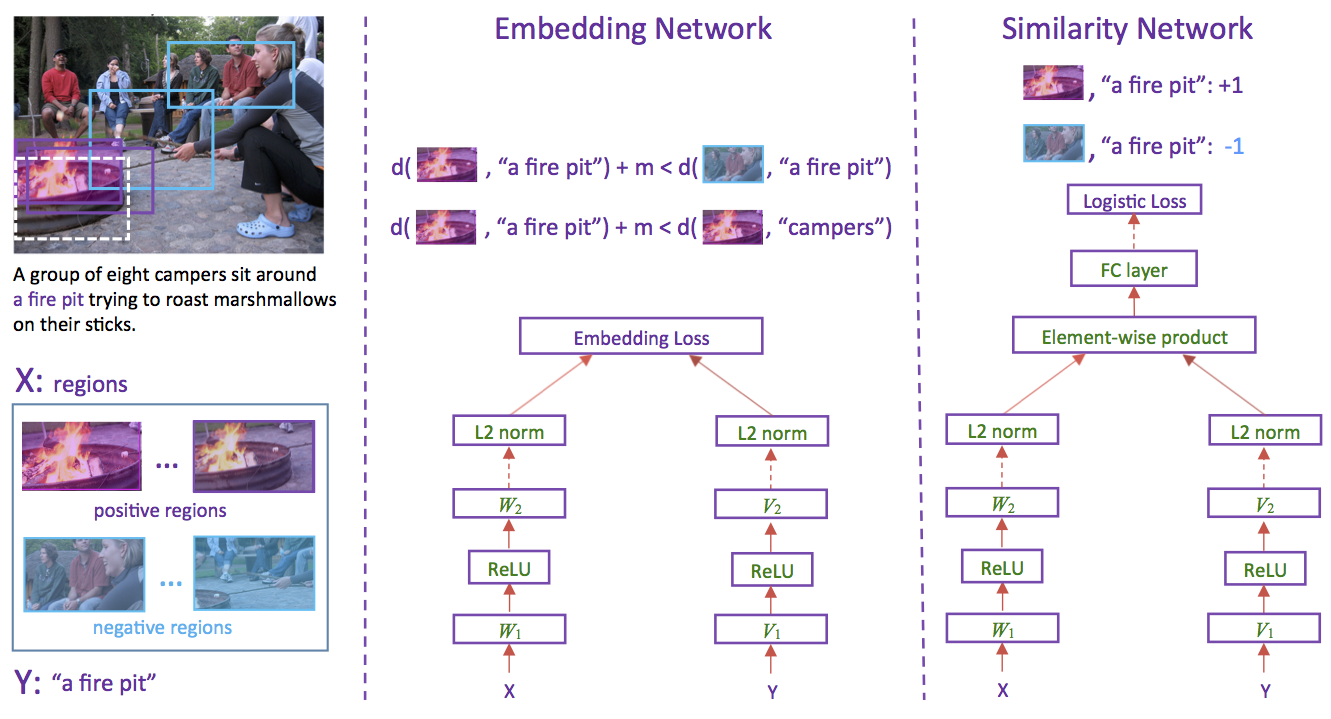}
\vspace{-0.5cm}
\caption{Taking the phrase localization task as an example, we show the architectures of the two-branch networks used in this paper. Left column: given the phrase ``a fire pit'' from the image caption, sets of positive regions (purple) and negative regions (blue) are extracted from the training image. The positive regions are defined as ones that have a sufficiently high overlap with the ground truth (dashed white rectangle). $X$ and $Y$ denote the feature vectors describing image regions and phrases, respectively. In this paper, $X$ are  features extracted from pre-trained VGG networks, and $Y$ are orderless Fisher Vector text features~\cite{klein2014fisher}. Middle: the embedding network. Each branch consists of fully connected (FC) layers with ReLU nonlinearities between them, followed by L2 normalization at the end. We train this network with a maximum-margin triplet ranking loss that pushes positive pairs closer to each other and negative pairs farther (Section \ref{sec:embedding}).
Right: the similarity network. As in the embedding network, the branches consist of two fully connected layers followed by L2 normalization. Element-wise product is used to aggregate features from two branches, followed by several additional fully connected (FC) layers. The similarity network is trained with the logistic regression loss function, with positive and negative image-text pairs receiving labels of ``+1'' and ``-1'' respectively (Section \ref{sec:similarity}).}
\label{fig:Model_structure}
\end{figure*}

We study neural architectures for a core problem underlying most image-text tasks---{\it how to measure the semantic similarity between visual data, e.g., images or regions, and text data, e.g., sentences or phrases}. Learning this similarity requires connecting low-level pixel values and high-level language descriptions. Figure~\ref{fig:Model_structure} shows an example of a phrase description of an image region from the Flick30K Entities dataset. Matching the phrase ``fire pit'' to its corresponding region requires not only distinguishing between the correct region and background clutter, but also understanding the difference between ``fire pit'' and other visual concepts that might be present in the image. Naively, one might consider training binary or multi-class classifiers to estimate the probabilities of various concepts given image regions. However, the natural language vocabulary of visual concepts is very large, even if we restrict these concepts to nouns or simple noun phrases. Further, different concepts have complex semantic similarity relationships between them -- for example, ``fire'' and ``flame'' are synonyms, ``fireplace'' is similar in meaning but not identical to ``fire pit,'' and attributes can modify the meaning of head nouns (``fire pit'' is not the same as ``pit''). This suggests that, instead of representing different phrases using separate classifiers, representing text in a continuous ``semantic'' embedding space is more appropriate. Furthermore, the frequencies of different phrases are highly unbalanced: the word ``fire'' only occurs three times in the Flickr30K Entities dataset, while the most common words, such as ``man,'' show up a few hundred times. For all these reasons, training separate per-concept classifiers is undesirable. A more natural approach is to design a model that takes in continuous image and text features (for the latter, derived from continuous word embeddings like word2vec~\cite{mikolov2013distributed}) and predicts a similarity score. This approach has the advantage of treating image and text symmetrically, enabling both image-to-text and text-to-image retrieval, and of being easily extendable from individual words and simple phrases to arbitrarily complex sentences, provided a continuous feature encoding for sentences can be devised.

As suggested by the above discussion, the network architecture for these tasks should consist of two branches that take in image and text features respectively, pass them through one or more layers of transformations, fuse them, and eventually output a learned similarity score. At a conceptual level, there are two ways to obtain this score. One is to train the network to map images and text into an explicit joint embedding space in which corresponding samples from the two modalities lie close to one another, and image-text similarity is given by cosine similarity or Euclidean distance. The second approach is to frame image/text correspondence as a binary classification problem: given an image/text pair, the goal is to output the probability that the two items match. Accordingly, we propose two variants of two-branch networks that follow these two strategies (Figure \ref{fig:Model_structure}): the {\em embedding network} and the {\em similarity network}. 


\textbf{Embedding Network}: The goal of this network is to map image and text features, which may initially have different dimensions, to a joint latent space of common dimensionality in which matching image and text features have high cosine similarity. Each branch passes the data through two layers with nonlinearities, followed by L2 normalization, so that cosine similarity is equivalent to Euclidean distance. We train the network with a bi-directional ranking loss that enforces that matched sample pairs should have smaller distance than unmatched ones in the embedding space. We also propose augmenting this loss with neighborhood information in each modality via novel triplet constraints and sampling strategy. In particular, where different phrases or sentences can be used to describe the same image or region, we force them to be close to each other. We argue that adding these constraints can help to regularize the learning of the embedding space, especially facilitating matching within the same modality, i.e., sentence-to-sentence retrieval.

\textbf{Similarity Network}: In our alternative architecture, image and text data is also passed through branches with two layers with nonlinearities, but then, element-wise product is used to aggregate features from the two branches into a single vector, followed by a further series of fully connected layers. This network is trained with logistic regression loss to match the output score to $+1$ for positive pairs and $-1$ for negative pairs.  
This network is notably simpler but also less flexible than our embedding network, as it no longer has an explicit embedding space and cannot encode structural constraints. However, it still achieves comparable performance on the phrase localization task.

Our contributions can be summarized as follows:
\begin{itemize}
\item We propose state-of-the-art {\em embedding} and {\em similarity networks} for learning the correspondence between image and text data for two tasks: {\em phrase localization} (given a phrase, find a corresponding bounding box in the image) and {\em bi-directional image-sentence search} (given a query image/sentence, retrieve matching sentences/images from a database). 
  \item We systematically investigate all important components of both embedding and similarity networks, including loss functions, feature fusion strategies, and different ways of sampling positive and negative examples to form mini-batches during training. 
  \item For the embedding network, we show how to take advantage of within-modality neighborhood structure via additional loss terms and a novel triplet sampling strategy, which can improve the accuracy of within-modality retrieval as well.
  \item We obtain state-of-the-art accuracies on phrase localization on the Flickr30K Entities dataset~\cite{plummer2015flickr30k}, and near state-of-the-art accuracies on bi-directional image-sentence retrieval on Flickr30K~\cite{young2014image} and MSCOCO~\cite{lin2014microsoft} datasets.\footnote{In the last few months, following the conclusion of our experiments, we have become aware of some new methods that achieve better Recall@1~\cite{eisenschtat2016linking}.}
\end{itemize}

A previous version of our embedding network was published in CVPR 2016~\cite{wang2015learning}. In this journal paper, we add the similarity network as a counterpoint to the embedding network, conduct more in-depth experiments, and significantly improve the absolute accuracies on both tasks. Our two-branch networks are very general and can be modified and applied to other image-text tasks. For example, our embedding architecture has already been used by~\cite{yu2016joint} for modeling referring expressions.

Section 2 covers the related work on learning from visual and text data. Section 3 presents our embedding and similarity networks. Sections 4 and 5 describe our experiments on phrase localization and image-sentence retrieval. Finally, Section 6 concludes with a summary of findings and a discussion of important future research directions.

\section{Related Work} \label{sec:related}

\noindent \textbf{CCA-based methods}. One of the most popular baselines for image-text embedding is Canonical Correlation Analysis (CCA), which finds linear projections that maximize the correlation between projected vectors from the two views~\cite{hardoon2004canonical,hotelling1936}. Recent works using it include~\cite{gong2014multi,gong2014improving,klein2014fisher}. To obtain a nonlinear embedding, other works have opted for kernel CCA~\cite{hardoon2004canonical,hodosh2013framing}, which finds maximally correlated projections in reproducing kernel Hilbert spaces with corresponding kernels. Despite being a classic textbook method, CCA has turned out to be a surprisingly powerful baseline. Klein et al.~\cite{klein2014fisher} showed that properly normalized CCA~\cite{gong2014multi} with state-of-the-art image and text features can outperform much more complicated models. The main disadvantage of CCA is its high memory cost, as it requires loading all the data into memory to compute the data covariance matrix. 
\smallskip

\noindent \textbf{Deep multimodal representations.}
To extend CCA to learning nonlinear projections and improve its scalability to large training sets, Andrew et al.\ \cite{andrew2013deep} and Yan and Mikolajczyk~\cite{mikolajczyk2015deep} proposed to cast CCA into a deep learning framework. Their methods are trained using stochastic gradient descent (SGD) and thus can be applied to large-scale datasets. However, as pointed out in~\cite{ma2015finding}, SGD cannot guarantee a good solution to the generalized eigenvalue problem at the heart of CCA  because covariance estimation in each minibatch is unstable. Our proposed networks share a similar two-branch architecture with deep CCA models~\cite{andrew2013deep,mikolajczyk2015deep}, but they are much more stable and accurate.

Apart from deep CCA, many other deep learning methods have been proposed for joint modeling of multiple modalities. Some of the earlier techniques have included restricted Boltzmann machines and autoencoders~\cite{ngiam2011multimodal,srivastava2012multimodal}. 
For image-text tasks, recurrent text representations are the most popular among current approaches~\cite{donahue2014long,kiros2014multimodal,kiros2014unifying,mao2014deep,venugopalan2014translating}. Unlike these works, we rely primarily on hand-crafted orderless text features from~\cite{klein2014fisher}. Our experiments of Section \ref{sec:image_sen} will show that these features perform similarly to LSTMs for image-sentence matching, which seems to suggest that bag-of-words information is sufficient for cross-modal tasks that do not involve generation of novel text, at least given the current state of image-text models. 
\smallskip

\noindent \textbf{Ranking-based methods}. Some of the most successful multi-modal methods, whether they be linear models or deep networks, are trained with a ranking loss. For example, WSABIE~\cite{weston2011wsabie} and DeVISE~\cite{frome2013devise} learn linear transformations of visual and text features into a shared space using a {\em single-directional} ranking loss, which applies a margin-based penalty to an incorrect annotation when it gets ranked higher than a correct one for describing an image. A {\em bi-directional} ranking loss adds the missing link in the opposite direction: It further ensures that for each annotation, the corresponding image gets ranked higher than unrelated images~\cite{karpathy2014deep,kiros2014unifying,socher2014grounded}. Our embedding network is also trained using bi-directional loss, but we carefully explore a number of implementation choices, resulting in a model that can significantly outperform CCA-based methods and scale to large datasets.
\smallskip

\noindent \textbf{Metric learning and Siamese networks.}
In our conference paper introducing the embedding network~\cite{wang2015learning}, in addition to the bi-directional ranking loss, we proposed constraints that preserve neighborhood structure within each individual view. Specifically, in the learned latent space, we want images (resp. sentences) with similar meaning to be close to each other. Such within-view neighborhood preservation constraints have been extensively explored in the metric learning literature~\cite{hu2014discriminative,mensink2012metric,shaw2011learning,shaw2009structure,weinberger2005distance,vzbontar2014computing}. In particular, the Large Margin Nearest Neighbor (LMNN) approach~\cite{weinberger2005distance} tries to ensure that for each image its target neighbors from the same class are closer than samples from other classes. As our work will show, these constraints are also helpful for the cross-view matching task, and for training models that can achieve high accuracy both for cross-view and within-view matching.

Our two-branch networks are related to Siamese networks for metric learning~\cite{bromley1993signature,chopra2005learning,han2015matchnet,hoffer2014deep,schroff2015facenet,wang2014learning}. However, instead of learning a similarity function between two instances from the same modality using tied weights, we learn the embedding space across two different modalities with asymmetric branches.

\noindent \textbf{Classification-based methods.} Learning the similarity between images and text can be also modeled as classification. Deep models can be designed to answer whether two input visual and text samples match each other~\cite{jabri2016revisiting,ba2015predicting,fukui2016multimodal}. For example, Jabri et al.~\cite{jabri2016revisiting} used a softmax function to predict whether the input image and question match with the answer choice for VQA. Ba et al.~\cite{ba2015predicting} trained a two-branch network using classification loss to match visual and text data for zero-shot learning. Rohrbach et al.~\cite{rohrbach2015grounding} used a softmax function to estimate the posterior probability of a phrase over all the available region proposals in an image. To fuse region and phrase features, they performed a linear transformation in each branch, followed by sum, followed by a ReLU nonlinearity and a fully connected (FC) layer. In a subsequent work, Fukui et al.~\cite{fukui2016multimodal} systematically investigated multiple feature fusion strategies and found element-wise product to be among the most effective. They then proposed a novel Multimodal Compact Bilinear Pooling (MCB) approach that slightly outperformed element-wise product. However, MCB has a high memory cost, necessitating the use of sketch approximations. Like~\cite{rohrbach2015grounding}, MCB uses softmax to map a phrase to a the single best region proposal from the image, with all the other regions (including ones having a high overlap with the ground truth) designated as negatives.

Our second network type, the similarity network, also builds on the idea of directly predicting similarity between a phrase and a region through classification. However, instead of softmax loss, we use non-exclusive logistic regression loss and treat each phrase-region pair as an independent binary classification problem -- that is, for a given phrase, more than one region in the same image can be positive. At training time, this allows us to augment the ground truth region for a phrase with other positive examples having a high overlap with it. As our experiments in Section \ref{sec_rp} will show, this positive data augmentation strategy plays a much more important role in improving performance than the fusion strategy, allowing us to outperform MCB using much simpler element-wise product.



\section{Embedding and Similarity Networks}

\subsection{Overview of Image-Text Tasks} \label{sec:tasks}

In this paper, we focus on two image-text tasks: phrase localization and image-sentence retrieval.
{\em Phrase localization}~\cite{plummer2015flickr30k}, also known as text-to-image reference resolution or visual grounding, has recently received lots of attention~\cite{wang2015learning,plummer2015flickr30k,fukui2016multimodal}. Our definition of phrase localization follows~\cite{plummer2015flickr30k}: given an image and an entity mention, i.e.\, noun phrase, taken from a sentence description that goes with that image, the goal is to predict the corresponding bounding box. We solve this task in a retrieval framework: Given the entity mention, we rank a few hundred candidate regions output by a separate region proposal method (e.g., EdgeBox~\cite{zitnick2014edge}) using the similarity score produced by one of our trained networks. Our embedding network computes cosine similarity scores between input phrase and candidate regions in the shared embedding space, while our similarity network directly output similarity scores via regression. 

Our second task, bi-directional {\em image-sentence retrieval}, refers both to image-to-sentence and sentence-to-image search. The definitions of the two scenarios are straightforward: given an input image (resp. sentence), the goal is to find the best matching sentences (resp. images) from a database. Both scenarios are handled identically by nearest neighbor search in the latent image-sentence embedding space. Our embedding network is the most appropriate for this task, as it directly optimizes the bi-directional ranking loss.


Sections \ref{sec:embedding} and \ref{sec:similarity} will explain the details of our two networks, their objective functions, and training procedures.

In the following, $X$ and $Y$ will denote the collections of training images and sentences or training regions and phrases, each encoded according to their own feature representation, and $x \in X$ and $y \in Y$ will denote individual image and text features. From now on, unless stated otherwise, when we use the term ``image and text,'' it applies equally to ``image and sentence'' or ``region and phrase.'' 

\subsection{Embedding Network}
\label{sec:embedding}

\subsubsection{Network Architecture}
The embedding network, illustrated in Figure \ref{fig:Model_structure} (middle), has two branches, each composed of a series of fully connected (FC) layers, separated by Rectified Linear Unit (ReLU) nonlinearities. We apply batch normalization~\cite{ioffe2015batch} right after the last FC layer (without ReLU) to improve the convergence during training. The output vectors are further normalized by their $L2$ norm for efficient computation of Euclidean distance. 

The embedding architecture is highly flexible. The two branches can have different numbers of layers. The inputs can be either pre-computed features or outputs of other networks (e.g. CNNs or RNNs), and back-propagation of gradients to the input networks is possible. In our work, we focus on investigating the behavior of the two-branch networks and thus stick to pre-computed image and text features, which already give us state-of-the-art results. 

\subsubsection{Learning Cross-Modal Matching by Ranking}

The embedding network is trained using stochastic gradient descent with a margin-based loss that encodes both bi-directional ranking constraints and neighborhood-preserving constraints within each modality. This section will discuss the design of our loss function and the strategy of sampling triplets for stochastic gradient descent. 
\smallskip

\noindent \textbf{Bi-directional ranking loss.} 
Given a visual input $x_i$ (a whole image or a region), let $Y^{+}_{i}$ and $Y^{-}_{i}$ denote its sets of matching (positive) and non-matching (negative) text samples, respectively. If $y_j$ and $y_k$ are positive and negative samples for $x_i$, we want the distance between $x_i$ and $y_j$ to be smaller than the distance between $x_i$ and $y_k$, with a margin of $m$. This leads to the following triplet-wise constraint:
\begin{equation} \label{eq:img2sen}
\begin{split}
d(x_i,  y_j) + m & < d(x_i, y_k) \\
 \forall y_j \in Y^{+}_i, \quad & \forall y_k \in Y^{-}_{i}.
\end{split}
\end{equation}
Note that here and in the following, $d(x,y)$ will denote the Euclidean distance between image and text features in the embedding space.

Given a text input $y_{i'}$ (a phrase or sentence), we have analogous constraints in the other direction: 
\begin{equation} \label{eq:sen2img}
\begin{split}
d(x_{j'}, y_{i'}) + m & < d(x_{k'}, y_{i'}) \\ 
\forall x_{j'} \in X^{+}_{i'}, \quad & \forall x_{k'} \in X^{-}_{i'},
\end{split}
\end{equation}
where $X^{+}_{i'}$ and $X^{-}_{i'}$ denote the sets of matched (positive) and non-matched (negative) visual data for $y_{i'}$.

These ranking constraints can be converted into a margin-based loss function:
\begin{equation} \label{eq:obj_simple}
\small
\begin{split}
&L(X, Y)=\lambda_{1} \sum_{i,j,k} [m + d(x_i,y_j) - d(x_i, y_k)]_+\\
& +\lambda_{2} \sum_{i',j',k'} [m + d(x_{j'}, y_{i'}) - d(x_{k'},y_{i'})]_+,\\
\end{split}
\end{equation}
where $[t]_+=max(0,t)$. Our bi-directional ranking loss sums over all triplets (a target instance, a positive match, and a negative match) defined in constraints (\ref{eq:img2sen}) and (\ref{eq:sen2img}).For simplicity, we fix the margin $m=0.05$ for all terms in our image-sentence experiments.
The weights $\lambda_1$ and $\lambda_2$ balance the strength of the ranking loss in each direction.


Optimizing the loss function requires enumerating triplets, which can be computationally expensive, especially for large datasets. Similarly to~\cite{karpathy2014deep,kiros2014unifying,socher2014grounded}, we use SGD to optimize the loss function and sample triplets within each mini-batch and our sampling strategy is loosely inspired by~\cite{joachims2009cutting,shaw2011learning}. Briefly, for each positive image-text pair $(x,y)$ in a mini-batch, we keep sampling triplets $(x,y,y')$ such that $(x,y')$ is a negative pair and $(y,x,x')$ such that $(x',y)$ is a negative pair. Details of triplet sampling algorithms for the phrase localization and image-sentence retrieval tasks will be given in Sections \ref{sec:detail} and \ref{sec:image_sen_minibatch}.   

\subsubsection{Preserving Neighborhood Structure within Modalities} \label{sec:neighborhood}

The many-to-many nature of correspondence for image-text tasks creates an additional aspect of complexity for training. For example, the same image region can be described by different phrases, while the same phrase can refer to different regions across the training set. These correspondences, in turn, induce neighborhood structure {\em within} each modality --- which text (resp. image) pairs are similar because they correspond to the same image (resp. text) example. It is therefore interesting to see how this structure can help in learning the embedding. 
\smallskip


\noindent \textbf{Neighborhood-preserving constraints.} In our conference paper~\cite{wang2015learning}, we proposed adding ``structure constraints'' (now termed neighborhood constraints) to our loss function. Let $N(x_{{i}})$ denote the neighborhood of $x_{{i}}$, which is the set of images or regions described by the same text as $x_{{i}}$. We would like to enforce a small margin of $m$ between $N(x_{{i}})$ and any data point $x$ outside of the neighborhood:
\begin{equation} \label{eq:structure_x}
\begin{split}
d(x_{{i}},x_{{j}}) + m & < d(x_{{i}}, x_{{k}}) \\
\forall x_{{j}} \in N(x_{{i}}), \quad & \forall x_{{k}} \not\in N(x_{{i}}),
\end{split}
\end{equation}

Analogously to (\ref{eq:structure_x}), we also define the neighborhood constraints for the text side:
\begin{equation} \label{eq:structure_y}
\begin{split}
d(y_{{i}'}, y_{{j}'}) + m & < d(y_{{i}'}, y_{{k}'}) \\
\quad \forall y_{{j}'} \in N(y_{i'}), \quad & \forall y_{{k}'} \not\in N(y_{{i}'}),
\end{split}
\end{equation}
where $N(y_{{i}'})$ is the set of descriptions, e.g. phrases or sentences, for the same visual data.

We then add terms corresponding to the above constraints to our baseline bi-directional ranking loss function in Eq.\ (\ref{eq:obj_simple}):
\begin{equation}\label{eq:obj}
\small
\begin{split}
&L_{st}(X, Y)=\lambda_{1} \sum_{i,j,k} [m + d(x_i, y_j) - d(x_i,y_k)]_+\\
& +\lambda_{2} \sum_{i',j',k'} [m + d(x_{j'}, y_{i'}) - d(x_{k'}, y_{i'}]_+,\\
& + \lambda_{3} \sum_{{i},{j},{k}} [m + d(x_{i},x_{j}) - d(x_{i}, x_{k})]_+  \\
& + \lambda_4 \sum_{{i}',{j}',{k}'} [m + d(y_{i'}, y_{j'}) - d(y_{i'}, y_{k'})]_+,
\end{split} 
\end{equation}
where the sums are over all triplets defined in the constraints (\ref{eq:img2sen}-\ref{eq:sen2img}) and (\ref{eq:structure_x}-\ref{eq:structure_y}). The weights $\lambda_3$ and $\lambda_4$ control the regularization power of the neighborhood-preserving terms, and small values give the best performance. 

For phrase localization, we typically have multiple phrases corresponding to the same region (derived from multiple sentences corresponding to the same image), and multiple regions corresponding to the same phrase (these can be regions in different training images, or overlapping positive regions in the same image). Thus, both neighborhood-preserving terms given by Eqs. (\ref{eq:structure_x}) and (\ref{eq:structure_y}) are meaningful. However, for image-sentence retrieval, while each image is paired with multiple sentences, Flickr30K and MSCOCO datasets do not allow us to determine when the same sentence can apply to multiple images. Therefore, the image-view constraints (Eq. \ref{eq:structure_x}) cannot be applied.
\smallskip

\noindent \textbf{Neighborhood sampling}. 
The use of neighborhood-preserving constraints requires that the same mini-batch contain more than one positive match for each target sample, i.e., at least two texts that are matched to the same image and vice versa. To ensure this, after  performing regular triplet sampling, for any target image feature $x$, we add new triplets to the mini-batch as necessary to ensure that there are at least two triplets $(x, y_1, y_1')$ and $(x, y_2, y_2')$ that pair the target $x$ with different positive matches $y_1$ and $y_2$ -- and analogously for any target text feature $y$. This is done by searching all positive pairs that contain $x$ (resp. $y$), which can be pre-computed using hash tables with small run-time cost (see Algorithm \ref{algo:minibatch} for details). 

In our original work~\cite{wang2015learning}, we introduced neighborhood sampling solely as a way to provide triplets upon which neighborhood constraints could be imposed. However, somewhat surprisingly, we have since found out that doing neighborhood sampling by itself, even without adding the corresponding terms to the objective function, already accounts for some improvements in image-sentence retrieval tasks. Accordingly, in Sections 4 and 5, we will evaluate the impact of our neighborhood sampling strategy apart from the constraints given by Eqs. (\ref{eq:structure_x}) and (\ref{eq:structure_y}).

\subsection{Similarity Network} \label{sec:similarity}

The complexity of the embedding network's objective function and triplet sampling strategy motivates us to consider as an alternative a more straightforward classification-based similarity network, shown on the right of Figure \ref{fig:Model_structure}. It shares the same architecture of the two branches as the embedding network, including FC, ReLU, batch normalization and $L2$ normalization. The network then merges the output of the two branches using element-wise product, followed by a series of FC and ReLU layers (we found three to give the best results). 

As discussed in Section \ref{sec:related}, our use of element-wise product for fusion of the two branches is inspired by the corresponding baseline of Fukui et al.~\cite{fukui2016multimodal}. Other methods of fusion, e.g., concatenation, bilinear pooling, and compact bilinear pooling, can also be used in our network in principle. However, as shown in~\cite{fukui2016multimodal}, element-wise product outperforms all other baseline strategies like concatenation, and typically comes within 1\% of the much more complex multimodal compact bilinear pooling method that is the main contribution of~\cite{fukui2016multimodal}. As our phrase localization experiments (Table \ref{phraselocal}) will show, even with element-wise product, our similarity network can still outperform the full model of~\cite{fukui2016multimodal}.

For each input pair $(x_i,y_j)$, the similarity network generates a score $p_{ij}$ seeking to match the correct ground truth label ($+1$ for positive pairs and $-1$ for negative pairs). Our training objective is thus a logistic regression loss defined over the samples $\{x_i, y_j, z_{ij}\}$, where $z_{ij}=+1$ if $x_i$ and $y_j$ match each other, and $-1$ otherwise:
\begin{equation}\label{eq:obj_sim}
\begin{split}
L(X, Y, Z) = & \sum_{i, j} \log (1+\exp(-z_{ij} p_{ij})) \,. \\
\end{split} 
\end{equation}

\noindent To train the similarity network with SGD, we only need to sample positive and negative image-text pairs, which is much simpler and more efficient than sampling triplets. The only subtlety we found is that it is necessary to balance the number of positive and negative pairs in each mini-batch. Otherwise, the network will be dominated by the large number of negative pairs. More specifically, we maintain an equal number of positives and negatives in every mini-batch, though the sizes of different mini-batches can vary, especially for the phrase localization task. More details of the sampling and training algorithms will be covered in Sections \ref{sec_rp} and \ref{sec:image_sen}.

\begin{Algorithm*}[t]
\begin{framed}
\noindent{\bf Triplet sampling for the embedding network.}
\begin{enumerate}
\item Accumulate all pairs of ground truth regions with corresponding phrases from the dataset. Randomly shuffle these pairs into sets of 100.
\item 
{\bf Positive region augmentation (optional):} For each ground truth region-phrase pair, generate a positive pair $(x,y)$ where $x$ is a randomly chosen region having IoU $\ge$ 0.7 with the ground truth region for phrase $y$. 
\item For each positive pair $(x, y)$ from the initial set of 100:
\begin{enumerate}
\item
Enumerate all triplets $(x, y, y')$ where $y'$ is a phrase from this mini-batch not in the same coreference chain as $y$ (i.e., phrases $y$ and $y'$ are not both associated with the same ground truth region). Evaluate the loss for all these triplets and keep at most $K=30$ with highest nonnegative loss.
\item
Enumerate all triplets $(y, x, x')$ where $x'$ is a region from the same image with IoU $<$ 0.3 with the ground truth. Keep at most $K=30$ triplets with highest nonnegative loss. \vspace{1ex}
\item
{\bf Neighborhood sampling (optional):}
\begin{enumerate}
\item
For each unique target region $x$: make sure there are at least two triplets $(x, y_1, y_1')$ and $(x, y_2, y_2')$ where $y_1$ and $y_2$ are both positive matches for $x$. If needed, add a triplet by sampling $y_2$ from the same coreference chain as $y_1$. 
\item
For each unique target phrase $y$: make sure there are at least two triplets $(y, x_1, x_1')$ and $(y, x_2, x_2')$ where $x_1$ and $x_2$ are both positive matches for $y$. If needed, add a triplet by randomly sampling a positive region $x_2$ from the same picture. 
\end{enumerate}
\end{enumerate}
\end{enumerate}
\noindent{\bf Pair sampling for the similarity network.}
\begin{enumerate}
\item Accumulate all pairs of ground truth regions with corresponding phrases from the dataset. Randomly shuffle these pairs into sets of 100.
\item {\bf Positive region augmentation (optional):} For each ground truth region-phrase pair, generate {\em all} positive pairs $(x,y)$ where $x$ is a region having IoU $\ge$ 0.7 with the ground truth region for phrase $y$. 
\item For each positive pair $(x,y)$ from the augmented set, get a corresponding  negative pair $(x',y)$ by randomly selecting a region proposal from the same image with IoU $<0.3$ with the ground truth. 
\end{enumerate}
\end{framed}
\vspace{-0.4cm}
\caption{Mini-batch construction for the phrase localization task.\label{algo:minibatch}}
\end{Algorithm*}

\section{Phrase Localization Experiments} 
\label{sec_rp}

This section presents our experiments on the task of phrase localization on the Flickr30K Entities benchmark~\cite{plummer2015flickr30k}. Flickr30K Entities augments the Flickr30K \cite{young2014image} image-sentence dataset, consisting of 31783 images with five sentences each, with annotations that link $244$K mentions of distinct entities in sentences to $276$K ground-truth bounding boxes. 

The phrase localization task was introduced in Section \ref{sec:tasks}. To recap briefly, given a query noun phrase from an image caption and a set of region proposals from the same image, we rank the proposals using the region-phrase similarity scores produced by one of our trained networks. Consistent with Plummer et al.~\cite{plummer2015flickr30k}, for each image we use 200 region proposals produced by the category-independent EdgeBox method~\cite{zitnick2014edge}. A  proposal is considered to be a correct match for the query phrase if it has an Intersection over Union (IoU) score of at least 0.5 with the ground-truth bounding box for that phrase. Accuracy is evaluated using Recall@K, defined as the percentage of phrases for which the correct region is ranked among the top K. 

\subsection{Training Set Construction}\label{sec:detail}

In our experience, properly defining positive/negative region-phrase pairs and sampling pairs and triplets of examples during training is crucial for achieving the best performance. Phrase localization is akin to object detection, in that region-phrase scores produced by the embedding should be sensitive not only to semantic correspondence, but also to localization quality, i.e., how much a given region proposal overlaps the ground truth box for a query phrase. By default, given a phrase from a description of a specific image, Flickr30K Entities annotations specify a unique ground truth region.\footnote{Consistent with \cite{plummer2015flickr30k}, for plural entities associated with multiple boxes, we form one big bounding box containing all the instances. We also exclude non-visual phrases, i.e., phrases that do not correspond to a bounding box.} Our conference paper~\cite{wang2015learning}, together with other related work like MCB~\cite{fukui2016multimodal}, only used the ground truth boxes as positive regions during training. However, we have since realized that it is highly beneficial to augment the ground truth positive region with other proposals that have sufficiently high overlap with it. Specifically, we consider proposals having $\text{IoU}>0.7$ with the ground truth as positive examples for the corresponding phrase, while proposals with $\text{IoU}<0.3$ are marked as negative background regions. As our experiments will demonstrate, positive region augmentation improves recall by $3\app 4\%$. It helps to improve the model's robustness since ground truth regions are not available at test time, and is consistent with the way object detection is typically evaluated (a detection does not need to perfectly overlap the ground truth box to be considered correct).  

Starting from our definition of positive and negative pairs, we sample mini-batches of triplets for the embedding network and pairs for the similarity network according to the procedure given in Algorithm \ref{algo:minibatch}.

\subsection{Baselines and Comparisons} \label{sec:rp_baselines}

Our experiments systematically evaluate multiple components of our models, including network structure, sampling of the training set, and different components of the loss function for the embedding network. The full list of variants used in our comparisons is as follows.
\smallskip

\noindent \textbf{Network Architecture.}
We are interested in how our networks benefit from being able to learn a nonlinear mapping in each branch. For this, we compare two variants:
\begin{itemize}
\item \textbf{Linear branch structure}: only keeping the first layers in each branch (i.e., the ones with parameters $W_{1}$, $V_{1}$, as shown in Figure \ref{fig:Model_structure}) immediately followed by $L2$ normalization.
\item \textbf{Nonlinear branch structure}: branches consisting of two FC layers with ReLU, batch normalization and $L2$ normalization. 
\end{itemize}

\noindent \textbf{Selecting Positive Pairs.} 
We evaluate how positive example augmentation contributes to the performance of phrase localization. We compare the vanilla scheme without augmentation to our scheme described in Section~\ref{sec:detail}:

\begin{itemize}
\item \textbf{Single positive}: using the ground truth region for a phrase as the single positive example.
\item \textbf{Augmented positive}: augmenting ground truth regions with other regions having IoU $>$ 0.7 with it.
\end{itemize}

\noindent \textbf{Embedding Loss Functions.}
In principle, phrase localization is a single-directional task of retrieving image regions given a query phrase, so we want to know whether we can derive an additional benefit by using a bi-directional loss function: 
\begin{itemize}
\item \textbf{Single-directional}: only using the phrase-to-region loss from Eq.(\ref{eq:obj}). This is done by setting $\lambda_1 = 0, \lambda_2 = 1, \lambda_3 = 0, \lambda_4 = 0$.  

\item \textbf{Bi-directional}: using the bi-directional loss from Eq.(\ref{eq:obj}). This is done by setting $\lambda_1 = 1, \lambda_2 = 4, \lambda_3 = 0, \lambda_4 = 0$. These parameter values have been tuned on our validation set.
\end{itemize}

\noindent \textbf{Neighborhood Sampling and Constraints.}
Flickr30K Entities dataset includes 130K pairs of region-phrase correspondences, with 70K unique phrases and 80K unique regions. In general, one phrase can correspond to many regions and vice versa. We are interested in how multiple matches to the same region (resp. phrase) can help the task, as described in Section \ref{sec:neighborhood}:

\begin{itemize}
\item \textbf{Neighborhood sampling}: using the sampling strategy of Section \ref{sec:neighborhood} to augment standard triplet sampling. 
\item \textbf{Neighborhood constraints}: using the full loss function of Eq.(\ref{eq:obj}). This is done by setting $\lambda_3 = 0.1,\lambda_4=0.1$. This requires the use of neighborhood sampling.
\end{itemize}

\subsection{Implementation details}

Following Rohrbach et al.~\cite{rohrbach2015grounding} and Plummer et al.~\cite{plummer2016ijcv}, we use Fast R-CNN features~\cite{girshick2015fast} from the VGG network~\cite{simonyan2014very} fine-tuned on the union of the PASCAL 2007 and 2012 train-val sets \cite{everingham2011pascal}. To be consistent with~\cite{plummer2016ijcv}, we extract $4096$D features from a single crop of an image region. 

For phrases, we use the Fisher Vector (FV) encoding~\cite{perronnin2010improving} as suggested by Klein et al.~\cite{klein2014fisher}. We start from $300$-dimensional word2vec features~\cite{mikolov2013distributed} and apply ICA as in~\cite{klein2014fisher} to construct a codebook with $30$ centers. The resulting FV representation has dimension $300\times 30\times 2 = 18000$. For simplicity, we only use the Hybrid Gaussian-Laplacian Mixture Model (HGLMM) from \cite{klein2014fisher} rather than the combined HGLMM+GMM model. To save memory and training time, we perform PCA on these $18000$-dimensional vectors to reduce the dimension to $6000$. A disadvantage of HGLMM is that it is a complex and nonlinear hand-crafted text feature. However, as we showed in the conference version of this work~\cite{wang2016structured}, we can obtain very similar results on top of basic tf-idf features. In this paper, our main focus is on the design and training of two-branch networks, so we omit the evaluation of different text features.

Both the embedding and similarity networks use the same configurations for the two branches. The image branch has two FC layers with weight matrices $W_1$ and $W_2$ having sizes $4096\times 1024$ and $1024\times 512$. The text branch has two FC layers with weight matrices $V_1$ and $V_2$ having sizes $6000\times 1024$ and $1024\times 512$. Thus, the embedding network projects image and text features into a 512-dimensional latent space. For the similarity network, the 512-dimensional outputs of the two branches get combined by the element-wise product layer that doesn't change the dimensionality, followed by three additional FC layers with parameters of size $512 \times 512$, $512 \times 256$ and $256 \times 1$. The similarity network outputs a scalar score trained with logistic regression as described in Section \ref{sec:similarity}. 

All our experiments in Sections \ref{sec_rp} and \ref{sec:image_sen} are conducted in TensorFlow. We train our phrase localization networks using the Adam optimizer~\cite{kingma2014adam} with an initial learning rate of 0.001. We use a mini-batch of $100$ image-phrase pairs. Each epoch thus needs around $4000$ iterations. Both our similarity (with single and augmented positives) and embedding networks (no neighborhood terms or sampling) converge after $32000$ iterations (around 8 epochs). For both similarity and embedding networks, it takes around 1-2 days to get convergence with a single Geforce GTX TITAN X card. Training the embedding network takes longer due to the addition of neighborhood constraints. Therefore, in order to save time, we first train the network without adding neighborhood constraints for $8$ epochs with neighborhood sampling, then add the neighborhood terms and resume the training for two additional epochs.

\subsection{Result Analysis}

At test time, we treat phrase localization as the task of retrieving regions matching a query phrase (assumed to be present in the image) from a set of region proposals. For the embedding network, the query phrase and each candidate region are passed through the respective branches to compute their embedded representation, and Euclidean distance (equivalently, cosine similarity) is used as the similarity score. For the similarity network, the score is predicted directly using the logistic formulation. In both cases, we rank regions in decreasing order of similarity to the query and report Recall@K, or the percentage of queries for which the correct match has rank of at most K. A region proposal is considered to be a correct match if it has IoU of at least 0.5 with the ground-truth bounding box for that phrase. We use the evaluation code provided by Plummer et al.~\cite{plummer2015flickr30k}. 

Table~\ref{phraselocal} shows the results of our embedding and similarity networks, in comparison to several state-of-the-art methods. Among them, CCA~\cite{plummer2015flickr30k}, GroundeR~\cite{rohrbach2015grounding} and MCB~\cite{fukui2016multimodal}, introduced in Section \ref{sec:related}, are representative linear and deep models for this task. We also compare to the structured matching system of Wang et al.~\cite{wang2016structured}, which uses a single-layer version of our embedding network from~\cite{wang2015learning} combined with global optimization to find a joint assignment of phrases to all image regions while satisfying certain relations derived from the sentence. As we can see from Table \ref{phraselocal}(a), CCA, which is trained only on positive region-phrase pairs, already establishes a strong baseline. MCB gives the best results among all previous methods. 

Table \ref{phraselocal}(b) gives the ablation study results for the embedding network. Using a nonlinearity within each branch improves performance, and using bi-directional instead of single-directional loss function doesn't give too much difference, since phrase localization emphasizes more on the single direction (phrase-to-region). Therefore, given the limited space, we only list those combinations with bi-directional loss in Table \ref{phraselocal}(b). Adding positive region augmentation improves $R@1$ by almost 5\%. Neighborhood sampling, that uses at least two positive regions for each query phrase in the mini-batch, gives a further increase of about $2\%$ over standard sampling that only uses one positive region. 

Finally, adding neighborhood constraints gives a slight drop in $R@1$ but further minor improvements in $R@5$ and $R@10$. Interestingly, contrary to our original expectations~\cite{wang2015learning}, it is the composition of the mini-batches, not the imposition of a specific neighborhood-preserving loss penalty during training, that seems to be responsible for most of the improvements in performance. However, image-sentence retrieval experiments in Section \ref{sec:image_sen} will demonstrate that neighborhood constraints have a more noticeable effect on the accuracy of retrieval within the same modality, i.e., sentence-to-sentence retrieval as opposed to image-sentence retrieval. For both neighborhood constraints and neighborhood sampling, as we found in experiments, the neighborhood of regions (namely the third term in Eq.(\ref{eq:obj})) plays a more important role than neighborhood of phrases (the fourth term in Eq.(\ref{eq:obj})) since it is easier to get multiple positive regions than have more than one positive phrases in this task.

Table \ref{phraselocal}(c) reports the accuracy of the similarity network with and without nonlinearity in each branch, with and without positive region augmentation. Consistent with the embedding network results, the nonlinear models improve $R@1$ over their linear versions by about $2\%$, but positive region augmentation gives an even bigger improvement of about $5\%$. The highest R@1 achieved by the similarity network is 51.05, which is almost identical to the 50.69 or 51.03 achieved by our best embedding networks. We also checked the performance of the similarity network with a different number of FC layers after the element-wise product, though we do not list the complete numbers in Table \ref{phraselocal} to avoid clutter. With a single FC layer, we get a significantly lower R@1 of 36.61, and with two FC layers, we get 49.39, which is almost on par with three layers.

\begin{figure*}
\begin{center}
\includegraphics[width=0.70\linewidth]{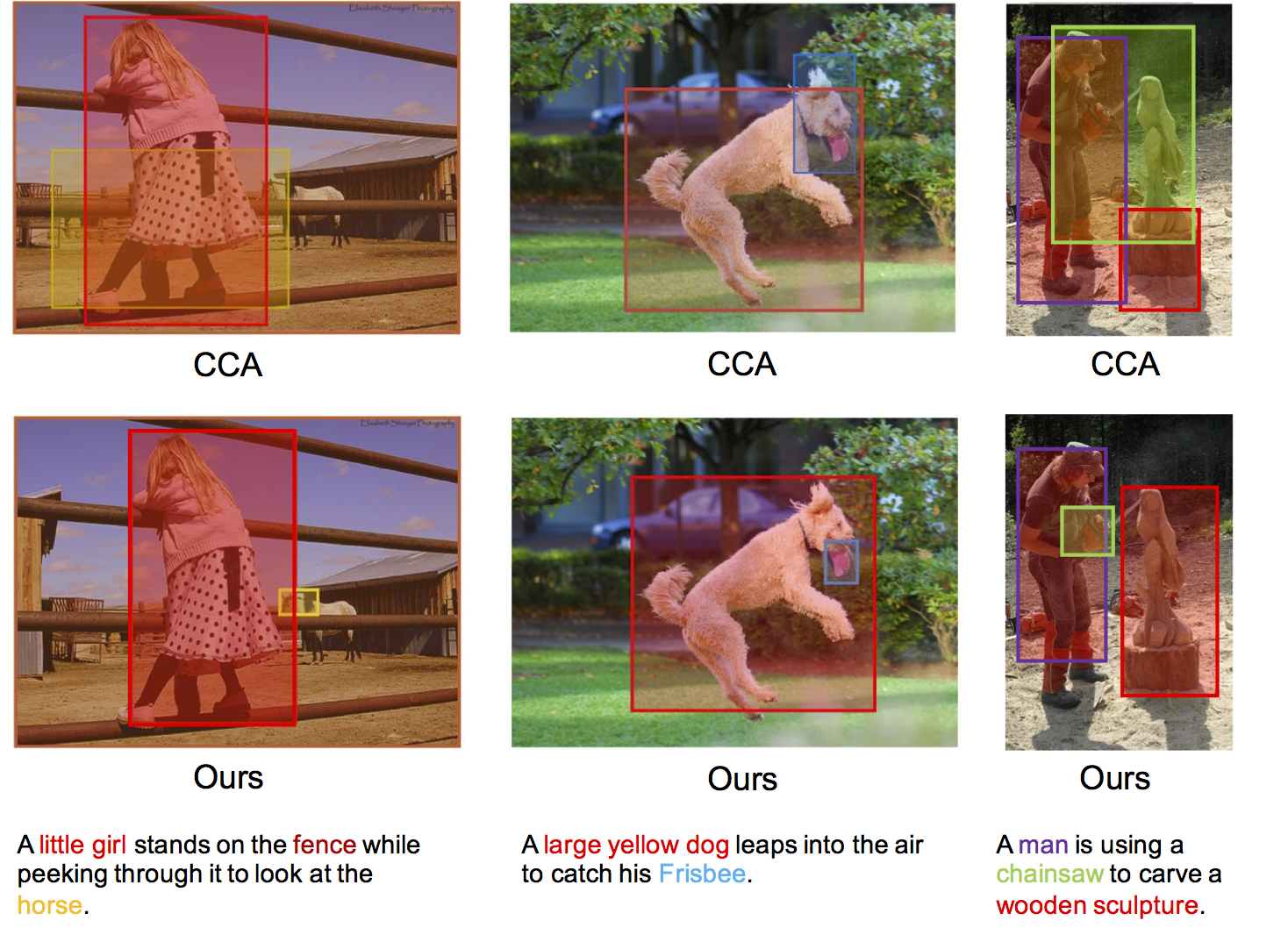}
\end{center}
\vspace{-0.3cm}
\caption{Example phrase localization results. For each image and reference sentence, phrases and best-scoring corresponding regions are shown in the same color. The first row shows the output of the CCA method~\cite{plummer2015flickr30k} and the second row shows the output of our best model (similarity network trained with augmented positive regions). In the first example, our method finds a partially correct bounding box for the horse while CCA completely misses it; in the second (middle) example, our method gives a more accurate bounding box for the frisbee. In the third (right) example, our method gives marginally better boxes for the chainsaw and wooden sculpture.}
\label{fig:exampleShow}
\end{figure*}

Figure \ref{fig:exampleShow} shows examples of phrase localization results in three images with our best model (similarity networks with augmented positives) compared to the CCA model.



\begin{table*}
\centering
\begin{tabular}{|l|l|l|l|l|l|l|l|l|l|l|l|}
\hline
                                                                                   \multirow{2}{*}{~} & \multicolumn{8}{l|}{\multirow{2}{*}{Methods on Flickr30K Entities}}   & 
                                                                                    \multirow{2}{*}{R@1}                  & \multirow{2}{*}{R@5}                  & \multirow{2}{*}{R@10}                  \\ 
 &          \multicolumn{8}{l|}{~} & & &                                                                          \\ \hline
\multirow{4}{*}{\begin{tabular}[c]{@{}l@{}}~\\(a)\\ State of\\ the art\end{tabular}}                                               & \multicolumn{8}{l|}{CCA baseline~\cite{plummer2015flickr30k}}                                                                                                                                                                                                                                                                                                                                                                                                                                                                                 & 41.77                & 64.52                & 70.77                 \\ 
                                                                                    & \multicolumn{8}{l|}{StructMatch~\cite{wang2016structured}}                                                                                                                                                                                                                                                                                                                                                                                                                                                                                     & 42.08                & -                    & -                     \\  
                                                                                    & \multicolumn{8}{l|}{GroundingPhrase~\cite{rohrbach2015grounding}}                                                                                                                                                                                                                                                                                                                                                                                                                                                                              & 47.70                & -                    & -                     \\  
                                                                                    & \multicolumn{8}{l|}{MCB Element-wise Product \cite{fukui2016multimodal}}                                                                                                                                                                                                                                                                                                                                                                                                                                                                             & 47.41                & -                    & -                     \\  
                                                                                    & \multicolumn{8}{l|}{MCB~\cite{fukui2016multimodal} }                                                                                                                                                                                                                                                                                                                                                                                                                                                                                                & 48.69                & -                    & -                     \\ \hline
\multirow{2}{*}{}                                                                   & 
\multirow{2}{*}{linear} & 
\multirow{2}{*}{\begin{tabular}[c]{@{}l@{}}non-\\ linear\end{tabular}} & 
\multirow{2}{*}{\begin{tabular}[c]{@{}l@{}}single\\ positive\end{tabular}} & \multirow{2}{*}{\begin{tabular}[c]{@{}l@{}}augm.\\ positive\end{tabular}} & 
\multirow{2}{*}{\begin{tabular}[c]{@{}l@{}}single-\\ dir.\end{tabular}} & \multirow{2}{*}{bi-dir.} &
\multirow{2}{*}{\begin{tabular}[c]{@{}l@{}}neighbor.\\ sampling\end{tabular}} & \multirow{2}{*}{\begin{tabular}[c]{@{}l@{}}neighbor.\\ constr.\end{tabular}} & 
\multirow{2}{*}{R@1} & \multirow{2}{*}{R@5} & \multirow{2}{*}{R@10} \\
                                                                                    &                                                                            &                                                                               &                                                                               &                                                                                  &                         &                            &                                                                             &                                                                         &                      &                      &                       \\ \hline
\multirow{6}{*}{\begin{tabular}[c]{@{}l@{}}(b)\\ Embedding\\ Network\end{tabular}} & \checkmark                                                                 & -                                                                             & \checkmark                                                                              & -                                                                                &   \checkmark                     & -                 &  -                                                                        & -                                                              & 39.60 & 64.30 & 71.00                                                        \\  

& -                                                             & \checkmark                                                                                 & \checkmark                                                                                 & -                                                                                 &   \checkmark                         & -                &     -                                                                      & -                                                              &  43.15        & 65.78               &  71.64                    \\ 

& -                                                                & \checkmark                                                                                 & \checkmark                                                                                & -                                                                                &   -                      & \checkmark                 &     -                                                                        & -                                                              & 44.53 &67.46 &73.12                     \\  

& -                                                                 & \checkmark                                                                             & -                                                                             & \checkmark                                                                                    &   -                      & \checkmark                 &     -                                                                        & -                                                              &  49.09 &69.46 &74.80                                      \\ 

                                                                                   & -                                                                          & \checkmark                                                                    & -                                                                   & \checkmark                                                                                   &   -                      & \checkmark                 &  \checkmark                                                                               & -                                                              & \textbf{51.03}     &  70.26     &  75.25                \\ 
                                                                                    & -                                                                          & \checkmark                                                                    & -                                                                    & \checkmark                                                                       &    -                     & \checkmark                 &  \checkmark                                                                           & \checkmark                                                              &  50.69          & \textbf{70.42}       & \textbf{75.51}     \\ \hline
\multirow{4}{*}{\begin{tabular}[c]{@{}l@{}}(c)\\ Similarity\\ Network\end{tabular}}  
& \checkmark                                                                 & -                                                                             & \checkmark                                                                                 & -                                                                                & -                     & -                 & -                                                                           & -                                                                       &  43.19              & 65.45               & 70.88                \\
& -                                                                & \checkmark                                                                                 & \checkmark                                                                             & -                                                                                & -                       & -                 & -                                                                           & -                                                                       & 45.19                & 67.14                & 72.13                 \\ 
                                                                                    &  \checkmark                                                                             & -                                                               & -                                                                             & \checkmark                                                                                    & -           & -                          & -                                                                           & -                                                                       & 48.30                   & 68.97                  & 74.08                    \\  
                                                                                    & -                                                                          & \checkmark                                                                    & -                                                                             & \checkmark                                                                                    & -                       & -                 & -                                                                           & -                                                                       & \textbf{51.05}     & 70.30                & 75.04                 \\ \hline
\end{tabular}
\vspace{0.1cm}
\caption{Phrase localization results on Flickr30K Entities.  We use 200 EdgeBox proposals, for which the recall upper bound is $R@200 = 84.58$. See Section \ref{sec:rp_baselines} for definitions of all the variants of embedding and similarity networks that we compare.}\label{phraselocal}
\end{table*}

\subsection{Discussion}
Our embedding and similarity networks are comparable in terms of recall, but they have different advantages and disadvantages. The embedding network has a more complex and memory-intensive training procedure due to its use of triplet sampling. To be specific, on average, the similarity network has 2138 pairs in each mini-batch (split equally between positive and negative pairs), while the embedding network has 5378 triplets without neighborhood sampling and 10756 with neighborhood sampling. On the other hand, the similarity network has more model parameters due to the addition of three FC layers after the element-wise product layer. The embedding network is more flexible than the similarity network because it learns an explicit ``shared space'' that is readily available for other tasks. The ranking loss used by the embedding network also gives us more freedom to fine-tune the structure of the embedding space via neighborhood-preserving constraints.


\section{Image-Sentence Retrieval} \label{sec:image_sen}

This section evaluates our networks on the task of bi-directional image-sentence retrieval, which was introduced in Section  \ref{sec:tasks}. Given a query image (resp. sentence), the goal is to find corresponding sentences (resp. images) from the dataset. In addition to the Flickr30K dataset, here we perform experiments on the larger MSCOCO dataset~\cite{chen2015microsoft}, consisting of 123287 images (the combination of released train2014 and val2014 from MSCOCO website) with five sentences each. MSCOCO does not include comprehensive region-phrase correspondence, so we can use it for image-sentence retrieval only.

\subsection{Training Set Construction} \label{sec:image_sen_minibatch}

The mini-batch construction procedure for the image-sentence task is a simplified version of the one from Algorithm \ref{algo:minibatch}. For the embedding network, we start by randomly permuting the data into mini-batches consisting of 500 positive image-sentence pairs. Then for each positive image-sentence pair $(x,y)$, we enumerate triplets $(x,y,y')$ where $y'$ is a sentence in the same mini-batch not associated with $x$, as well as triplets $(y,x,x')$ where $x'$ is an image in the mini-batch not associated with $y$. In both cases, we keep at most $K=10$ triplets with highest nonnegative loss. For neighborhood sampling, we need to make sure that given a target image $x$, a mini-batch has at least two triplets $(x,y_1,y_1')$ and $(x,y_2,y_2')$ where $y_1$ and $y_2$ are both sentences associated with $x$. Because we typically cannot identify more than one image described by the same sentence, we cannot do the other direction of neighborhood sampling, corresponding to step 3.c.ii of Algorithm \ref{algo:minibatch}.

For the similarity network, for each positive pair $(x,y)$, we generate a negative pair $(x,y')$ by randomly sampling a sentence not associated with the image. Note, however, that we cannot guarantee that $x$ and $y'$ are semantically incompatible with each other, since for any given image, our image-sentence datasets probably contain a number of sentences not associated with it that could still describe it accurately. This is not a major issue for phrase localization, since in Section \ref{sec_rp} we adopt the alternative strategy of sampling negative pairs $(x',y)$ with the ``negative'' regions $x'$ constrained to have low overlap with $x$ in the same image (Algorithm \ref{algo:minibatch}). It also doesn't play as much of a role for the embedding network, since the triplet objective merely tries to make sure that the captions actually written for a given image are closer to it than other sentences, not to push down the similarity of the other sentences to the image to a fixed low target value. Based on this reasoning, we expect the similarity network to perform poorly for image-sentence retrieval, and the subsequent experiments confirm this expectation.

\begin{table*}[t]
\centering
\begin{tabular}{|l|l|l|l|l|l|l|l|l|l|l|l|l|}
\hline
\multirow{2}{*}{}                     & \multicolumn{6}{l|}{\raisebox{-1.5ex}{Methods on Flickr30K}}                                                                                                                                                                                                                                                                                 & \multicolumn{3}{l|}{Image-to-sentence}              & \multicolumn{3}{l|}{Sentence-to-image}              \\ \cline{8-13} 
                                      & \multicolumn{6}{l|}{}                                                                                                                                                                                                                                                                                                     & R@1             & R@5             & R@10            & R@1             & R@5             & R@10            \\ \hline
\multirow{4}{*}{(a) State of the art} & \multicolumn{6}{l|}{Deep CCA~\cite{mikolajczyk2015deep}}                                                                                                                                                                                                                                                           & 27.9            & 56.9            & 68.2            & 26.8            & 52.9            & 66.9            \\ 
                                      & \multicolumn{6}{l|}{mCNN(ensemble)~\cite{ma2015multimodal}}                                                                                                                                                                                                                                                        & 33.6            & 64.1            & 74.9            & 26.2            & 56.3            & 69.6            \\ 
                                      & \multicolumn{6}{l|}{m-RNN-vgg~\cite{mao2014deep}}                                                                                                                                                                                                                                                                  & 35.4            & 63.8            & 73.7            & 22.8            & 50.7            & 63.1            \\ 
                                      & \multicolumn{6}{l|}{Mean vector~\cite{klein2014fisher}}                                                                                                                                                                                                                                                            & 24.8            & 52.5            & 64.3            & 20.5            & 46.3            & 59.3            \\ 
                                      & \multicolumn{6}{l|}{CCA (FV HGLMM)~\cite{klein2014fisher}}                                                                                                                                                                                                                                                         & 34.4            & 61.0            & 72.3            & 24.4            & 52.1            & 65.6            \\ 
                                      & \multicolumn{6}{l|}{CCA (FV GMM+HGLMM)~\cite{klein2014fisher}}                                                                                                                                                                                                                                                     & 35.0            & 62.0            & 73.8            & 25.0            & 52.7            & 66.0            \\ 
                                      & \multicolumn{6}{l|}{CCA (FV HGLMM)~\cite{plummer2015flickr30k}}                                                                                                                                                                                                                                                    & 36.5            & 62.2            & 73.3            & 24.7            & 53.4            & 66.8        \\ 
& \multicolumn{6}{l|}{Two-way Nets~\cite{eisenschtat2016linking}}                                                                                                                                                                                                                                                  & {49.8}           &   67.5       &    -        &  {36.0}         &   55.6         & -    \\ \hline
                                      & linear     & \begin{tabular}[c]{@{}l@{}}non-\\ linear\end{tabular} & \begin{tabular}[c]{@{}l@{}}single\\ dir.\end{tabular} & bi-dir. & \begin{tabular}[c]{@{}l@{}}neighbor.\\ sampling\end{tabular} & \begin{tabular}[c]{@{}l@{}}neighbor.\\ constr.\end{tabular} & R@1             & R@5             & R@10            & R@1             & R@5             & R@10            \\ \hline
\multirow{6}{*}{(b) Embedding Network}  & \checkmark &   -                                                    & \checkmark                                                 &       -                                                 &       -                                                       &     -                                                            & 38.1 &68.9 &79.4 & 22.6 & 49.0 &61.5			             \\ 
                                      & \checkmark &  -                                                     &   -                                                         &  \checkmark                                             &        -                                                      &      -                                                           &  40.1 &66.9 &75.9 & 27.5 &56.9 &68.5   \\ 
                                      & -           & \checkmark                                            & \checkmark                                                 &     -                                                   &     -                                                         &    -                                                             & 40.5 &71.9 &80.8 &22.1 &47.1 &57.9    \\ 
                                      & -           & \checkmark                                            &      -                                                      & \checkmark                                             &      -                                                        &     -                                                            &  40.5 &70.7 &80.9 &30.7 &61.1 &72.3			          \\ 
                                      &  -          &  \checkmark                                            &     -                                                       & \checkmark                                             & \checkmark                                                   &     -                                                            &   41.6 & 69.5 & 79.3 &31.1 &61.5 &72.8			              \\ 
                                      &  -          & \checkmark                                            &     -                                                       & \checkmark                                             & \checkmark                                                   & \checkmark                                                      & 43.2 &71.6 &79.8 &31.7 &61.3 &72.4			\\ \hline
\multirow{1}{*}{(c) Embedding Network(LSTM)}  & - & \checkmark & - & \checkmark & \checkmark & \checkmark & 37.5 & 64.7 &75.0 & 28.4 &56.3 &67.4 \\  \hline  
\multirow{1}{*}{(d) Similarity Network}  & - & \checkmark & - & - & - & - & 16.6  &   38.8   &  51.0
& 7.4   &   23.5  &  33.3 \\  \hline

\end{tabular}
\vspace{0.1cm}
\caption{Bi-directional retrieval results. The numbers in (a) come from published papers, and the numbers in (b-d) are results of our embedding and similarity networks. Note that the Deep CCA results in~\cite{mikolajczyk2015deep} were obtained with AlexNet~\cite{krizhevsky2012imagenet}. The results of our embedding network with AlexNet are still about 3\% higher than those of~\cite{mikolajczyk2015deep} for image-to-sentence retrieval and 1\% higher for sentence-to-image retrieval.}
\label{table:flickr30k}
\end{table*}

\subsection{Baselines and Comparisons}

Just as in Section \ref{sec_rp}, we demonstrate the impact of different components of our models by reporting results for the following variants.
\smallskip

\noindent \textbf{Linear vs. Nonlinear Branch Structure.} The same way as in the phrase localization experiments, we want to see the difference made by having one vs. two fully connected layers within each branch.
\smallskip 

\noindent \textbf{Embedding Loss Functions.}
Image-sentence retrieval is a bi-directional retrieval task, so we want to see whether bi-directional loss can give a bigger improvement over the single-directional loss than that on phrase localization task. 

\begin{itemize}
\item \textbf{Single-directional}: in Eq.(\ref{eq:obj}), only using the single direction (from image to sentences) by setting $\lambda_1 = 1, \lambda_2 = 0, \lambda_3 = 0, \lambda_4 = 0$.  

\item \textbf{Bi-directional}: in Eq.(\ref{eq:obj}), set $\lambda_1 = 1, \lambda_2 = 1.5, \lambda_3 = 0, \lambda_4 = 0$. These parameter values are determined on the validation set.
\end{itemize}

\noindent \textbf{Neighborhood Sampling and Constraints.}
In both Flickr30K and MSCOCO datasets, each image is associated with five sentences. Therefore, we can try to enforce neighborhood structure on the sentence space. We cannot do it on the image space since in the Flickr30K and MSCOCO datasets we do not have direct supervisory information about multiple images that can be described by the same sentence. Thus, in Eq.(\ref{eq:obj}), we always have $\lambda_3=0$.
\begin{itemize}
\item \textbf{Neighborhood sampling}: using the neighborhood sampling strategy (see Section \ref{sec:neighborhood}) to replace standard triplet sampling. 
\item \textbf{Neighborhood constraints}: using the full loss function as in Eq.(\ref{eq:obj}). This is done by setting $\lambda_2 = 1.5,\lambda_4=0.05$. We always use neighborhood sampling in this case. 
\end{itemize}

\begin{table*}[t]
\centering
\vspace{-0.5cm}
\begin{tabular}{|l|l|l|l|l|l|l|l|l|}
\hline
                                      & \multicolumn{2}{l|}{\raisebox{-1.5ex}{Methods on MSCOCO 1000}}                                                                         & \multicolumn{3}{c|}{Image-to-sentence}              & \multicolumn{3}{c|}{Sentence-to-image}              \\ \cline{4-9}
                                      & \multicolumn{2}{l|}{}                                                                                                           & R@1             & R@5             & R@10            & R@1             & R@5             & R@10            \\ \hline
\multirow{8}{*}{(a) State of the art} & \multicolumn{2}{l|}{Mean vector~\cite{klein2014fisher}}                                                                  & 33.2            & 61.8            & 75.1            & 24.2            & 56.4            & 72.4            \\
                                      & \multicolumn{2}{l|}{CCA (FV HGLMM)~\cite{klein2014fisher}}                                                               & 37.7            & 66.6            & 79.1            & 24.9            & 58.8            & 76.5            \\
                                      & \multicolumn{2}{l|}{CCA (FV GMM+HGLMM)~\cite{klein2014fisher}}                                                           & 39.4            & 67.9            & 80.9            & 25.1            & 59.8            & 76.6            \\
                                      & \multicolumn{2}{l|}{DVSA~\cite{karpathy2015deep}}                                                                    & 38.4            & 69.9            & 80.5            & 27.4            & 60.2            & 74.8            \\
                                      & \multicolumn{2}{l|}{m-RNN-vgg~\cite{mao2014deep}}                                                                        & 41.0            & 73.0            & 83.5            & 29.0            & 42.2            & 77.0            \\
                                      & \multicolumn{2}{l|}{mCNN(ensemble)~\cite{ma2015multimodal}}                                                              & 42.8            & 73.1            & 84.1            & 32.6            & 68.6            & 82.8            \\
                                      & \multicolumn{2}{l|}{LayerNorm~\cite{ba2016layer}}                                                                       & 48.5            & {80.6}            & {89.8}            & 38.9            & 74.3            & 86.3            \\
                                      & \multicolumn{2}{l|}{OrderEmbedding ~\cite{vendrov2015order}}                                                             & 46.7            & -               & 88.9            & 37.9            & -               & 85.9            \\ 
                                      
& \multicolumn{2}{l|}{Two-way Nets~\cite{eisenschtat2016linking}}                                                                       &  {55.8}         &  75.2        &    -        &  39.7         &   63.3        & -    \\ \hline
                                      & \begin{tabular}[c]{@{}l@{}}neighbor.\\ sampling\end{tabular} & \begin{tabular}[c]{@{}l@{}}neighbor. \\ constr.\end{tabular} & R@1             & R@5             & R@10            & R@1             & R@5             & R@10            \\ \hline
 \multirow{3}{*}{\begin{tabular}[c]{@{}l@{}} (b) Embedding Network \\ (nonlinear, bi-directional)\end{tabular}}   & -                                                            & -                                                                &    53.0		         &   82.3        &     91.6       & 41.2            &    75.8         &  87.7           \\ 
                                      & \checkmark                                                   & -                                                                &  54.2         &  84.1    &   92.2   &   42.4       &   76.2    &     88.1       \\ 
                                      & \checkmark                                                   & \checkmark                                                       & 54.9    &      84.0       &      92.2      &  43.3  &   76.4  & 87.5   \\ \hline
\multirow{1}{*}{(c) Embedding  Network(LSTM)} & \checkmark & \checkmark & 54.0 & 84.0 & 91.2 &43.3 &76.8 & 87.6  \\ \hline
\multirow{1}{*}{(d) Similarity  Network} & - & - & 30.9  & 61.1  &   76.2 & 14.0   &  30.0  &  37.8 \\ \hline
\end{tabular}
\vspace{0.1cm}
\caption{Bi-directional retrieval results on the MSCOCO 1000-image test set.}
\label{table:microsoftCOCO1000_table}
\end{table*}

\subsection{Implementation Details}

To represent whole images, we follow the implementation details in \cite{klein2014fisher,plummer2015flickr30k}. Given an image, we extract the 4096-dimensional activations from the 19-layer ImageNet-trained VGG model~\cite{simonyan2014very}. Following standard procedure, the original $256\times256$ image is cropped in ten different ways into $224\times224$ images: the four corners, the center, and their x-axis mirror image. The mean intensity is then subtracted from each color channel, the resulting images are encoded by the network, and the network outputs are averaged. The output dimensions of the two FC layers on the image side are 2048 and 512. 

To represent sentences, we continue to rely on the same orderless HGLMM features as for phrase localization, PCA-reduced to 6000 dimensions, with output dimensions of the two FC layers on the text side also set to 2048 and 512. However, while we could be reasonably assured that these features do not lose much information when representing short phrases, their suitability for longer sentences is less clear. Therefore, in this section we also evaluate a recurrent sentence representation learned by an LSTM~\cite{hochreiter1997long}. We start with a one-hot encoding with vocabulary size of 11,263 (MSCOCO) and 8,569 (Flickr30K), which are the numbers of words in the respective training sets. This input gets projected into a word embedding layer of dimension 256, and the LSTM hidden space dimension is 512. The hidden space output is used as the input to the text branch of our embedding network. Accordingly, we change the first FC layer of the text branch to accept 512-dimensional input. The output dimensions of the two FC layers are 1024 and 512. During training, the LSTM parameters are optimized jointly with the rest of the network parameters by back-propagating the embedding loss.

For the subsequent experiments, we train our networks using Adam with starting learning rate of 0.0001 for HGLMM features and 0.0002 for LSTM features.
We use a Dropout layer after ReLU with probability = 0.5 (note that in the phrase localization experiments of Section \ref{sec_rp}, we did not use Dropout, as we found that it did not make a difference).


\subsection{Result Analysis}

For evaluation of bi-directional image-sentence retrieval, we follow the same protocols as other recent works~\cite{klein2014fisher,plummer2015flickr30k}. 
Given the test set of 1000 images and 5000 corresponding sentences, we use our networks to score images given query sentences and vice versa, and report performance as Recall@$K$ ($K=1,5,10$), or the percentage of queries for which at least one correct ground truth match was ranked among the top $K$ matches. 
For Flickr30K, we use the same random split as Plummer et al.~\cite{plummer2015flickr30k}.
For MSCOCO, like~\cite{karpathy2014deep,klein2014fisher}, we randomly generate the splits that contain 113287 images with their corresponding sentences for training, 1000 images and their corresponding sentences for testing and the remaining images and their corresponding sentences for validation. 

Results on the Flickr30K and MSCOCO datasets are given in Tables \ref{table:flickr30k} and \ref{table:microsoftCOCO1000_table}, respectively. Parts (a) of the tables list the numbers reported by recent competing methods. The most relevant baseline for our embedding network is CCA (HGLMM)~\cite{klein2014fisher,plummer2015flickr30k}, since it uses the same underlying feature representations for images and sentences. Parts (b) of the tables give results for our embedding networks, and the trends are largely similar to those of Table \ref{phraselocal}. Going from single-directional to bi-directional constraints improves the accuracy by a bigger amount for sentence-to-image retrieval. Neighborhood sampling is effective and can generally improve over conventional triplet sampling around in R@$1$ across the table, and adding neighborhood constraints does not show significant further improvements. In Table \ref{table:microsoftCOCO1000_table}(b), adding neighborhood constraints improves the R@1 in both directions but shows a small drop for R@10. However, we will show in Section \ref{sec:within_view} that adding neighborhood constraints can consistently improve within-view retrieval.


Parts (c) of Tables \ref{table:flickr30k} and \ref{table:microsoftCOCO1000_table} give the results for our full embedding network with LSTM sentence encoding, which turns out to be comparable to, or slightly worse than, the HGLMM feature. For completeness, Parts (d) give the results for the similarity network. While we argued in Section \ref{sec:image_sen_minibatch} that the similarity network is poorly suited for this task, it is remarkable just how low its numbers are, especially given its competitive accuracy on the phrase localization task.




\subsection{Sentence-to-sentence Retrieval} \label{sec:within_view}
Our experiments on the embedding network both for phrase localization and image-sentence retrieval have shown that neighborhood sampling can give considerable improvements even without adding neighborhood constraint terms to the triplet loss. Thus, it is still unclear how neighborhood constraints change the latent embedding space. Therefore, in this section, instead of only looking at cross-modal retrieval, we show how neighborhood constraints can improve performance for the within-view task of sentence-to-sentence retrieval: given a query sentence, retrieve other sentences that correspond to the same image. For the evaluation metric, we still use R@K. We also use the same training/val/testing splits as in the previous section. Results on Flickr30K and MSCOCO datasets are listed in Table \ref{sen2sen}. It can be seen that adding neighborhood constraints on top of neighborhood sampling provides a more convincing gain for within-view retrieval than for cross-view retrieval. This behavior can be useful for practical multi-media systems where both tasks are required at the same time. 

\begin{table}
\begin{center}
\begin{tabular}{|l|l|l|l|l|}
\hline
\multicolumn{5}{|l|}{\multirow{2}{*}{Methods on Flickr30K}} \\ 
\multicolumn{5}{|l|}{} \\ \hline               
\begin{tabular}[c]{@{}l@{}}neighbor.\\ sampling\end{tabular}                  & \begin{tabular}[c]{@{}l@{}}neighbor.\\ constr.\end{tabular}                  & R@1                   & R@5                  & R@10                  \\ \hline
-                                                                             & -                                                                                &60.5 & 81.4 &87.7	                 \\ 
                                                                \checkmark              & -                                                                                &60.5 &81.5 &87.6	             \\ 
                                                                   \checkmark           &    \checkmark                                                                              &  63.8 &84.1 &90.2   \\ \hline
\multicolumn{5}{|l|}{\multirow{2}{*}{Methods on MSCOCO}} \\   
\multicolumn{5}{|l|}{} \\ \hline             
\begin{tabular}[c]{@{}l@{}}neighbor.\\ sampling\end{tabular}                  & \begin{tabular}[c]{@{}l@{}}neighbor.\\ constr.\end{tabular}                  & R@1                   & R@5                  & R@10                  \\ \hline
-                                                                             &       -                                                                           &    54.4 &78.3 &86.9                       \\ 
                                                                    \checkmark         &    -                                                                              &   54.3 &78.8  & 86.9	                          \\ 
                                                                  \checkmark          &  \checkmark                                                                             & 55.5  & 79.6 &87.8	     \\ \hline
\end{tabular}
\vspace{0.1cm}
\caption{Sentence-to-sentence retrieval on Flickr30K and MSCOCO datasets.}
\label{sen2sen}
\end{center}
\end{table}

\begin{table*}[h]
 \centering
 \begin{tabular}{|l|l|l|l|l|l|l|}
 \hline
Methods on Flickr30K & \multicolumn{3}{l|}{Image-to-sentence} & \multicolumn{3}{l|}{Sentence-to-image} \\ \hline
 & R@1            & R@5            & R@10          & R@1            & R@5            & R@10          \\ \hline
Image-sentence model   & 43.2 &71.6 &79.8 &31.7 &61.3 &72.4    \\ 
 Weighted distance  & 43.8  & 72.1  & 80.4     & 33.5   & 62.4   & 73.9 \\ \hline
\end{tabular}
\vspace{0.1cm}
\caption{Results on Flickr30K image-sentence retrieval with incorporating region-phrase correspondences (see text).}
\label{table:weighted}
\end{table*}

\subsection{Combining Image-Sentence and Region-Phrase Models}

So far, we have evaluated our networks separately on region-phrase and image-sentence correspondence tasks. The next obvious question is whether the local and global similarity models can be brought together, for example, to improve performance on image-sentence retrieval. Intuitively, it would be nice to have an approach that can verify whether an image and a sentence match based not only on their global features, but on detailed correspondence between regions in the image and phrases in the sentence. Given the high accuracy achieved by our models on phrase localization in Section \ref{sec_rp}, one would expect that combining it with the image-sentence model of Section \ref{sec:image_sen} would lead to significant improvements. However, one of the most frustrating findings in our work to date is that making such a combination work is highly non-trivial. For completeness, and to point towards one of our most important future directions, we give in this section the results of the simple post-hoc weighted combination scheme from our related work~\cite{plummer2016ijcv}. Given an image $x$ and a sentence $y$, we define the combined image-sentence and region-phrase distance as 
\begin{equation}\label{eq:weighted}
D(x,y) = (1-\alpha)\,d(x,y) + \alpha \,d_{rp}(x,y) \,,
\end{equation}

\noindent where $d(x,y)$ is the distance in the image-sentence latent space learned by the embedding network, and $d_{rp}(x,y)$ is the average of the distances between all the phrases in the sentence and their best-matching regions in the image. For the image-sentence model, we use the best embedding network from Table~\ref{table:flickr30k}. For the region-phrase model, we use the generated embedding distance matrix from Table \ref{phraselocal}. We set $\alpha = 0.3$, which was experimentally found to give the best results. 

As can be seen from Table \ref{table:weighted}, the improvement of the combined model over the image-sentence one is very small. We have analyzed some of the reasons for this somewhat surprising and frustrating outcome in our related journal paper~\cite{plummer2016ijcv}, where we used simple CCA embeddings. Namely, the global image-sentence model already works very well for image-sentence retrieval, in that it usually succeeds in retrieving sentences that roughly fit the image. In order to provide an improvement, the region-phrase model would have to make fine distinctions of which it is currently incapable, e.g., between closely related or easily confused objects, between fine-grained attributes of people, or cardinalities of people and objects. Furthermore, due to the way our region-phrase model is trained, its scores are meant to be useful for ranking regions within the same image based on the correspondence to a given phrase that is assumed to be present. However, we found them to be much less consistent when ranking a phrase across different images, or ranking different phrases in the same image. In other words, given a region-phrase score output by our embedding network, we cannot use it as evidence of presence or absence of a given phrase in an image, in the same way that one might want to use the score of an object detector. Extending our training formulation to a more open-ended phrase detection scenario is an important subject for future work, as is joint training of image-sentence and region-phrase embeddings in a single network (to date, our attempts at joint training have not led to good results).


\section{Conclusion and Future Work}
This paper has studied state-of-the-art two-branch network architectures for region-to-phrase and image-to-sentence matching. To our knowledge, our results on Flickr30K and MSCOCO datasets are the best to date on both tasks. Our first architecture, the {\em embedding network}, works by explicitly learning a non-linear mapping from input image and text features into a joint latent space in which corresponding image and text features have high similarity. This network works well for both image-sentence and region-phrase tasks, though its objective consists of multiple terms and relies on somewhat costly and intricate triplet sampling. We investigated triplet sampling within mini-batches in detail and showed that the way it is done can have a significant impact on performance, even without changing the objective function. Our second architecture, the {\em similarity network}, tries to directly predict whether an input image and text feature are similar or dissimilar. Our experiments showed that this network can serve as an attractive alternative to the embedding network for region-phrase matching, but fails miserably for image-sentence retrieval, revealing an interesting difference between the two tasks. Finally, our preliminary unsuccessful experiments on combining image-sentence and region-phrase models indicate an important direction for future research.

\section*{Acknowledgments}
This material is based upon work supported by the National
Science Foundation under Grants CIF-1302438 and IIS-1563727, Xerox
UAC, and the Sloan Foundation. We would like to thank
Bryan Plummer for providing features for region-phrase experiments, and Kevin Shih for thoughtful discussions on the similarity network and help with building the region-phrase experimental environment.

\bibliographystyle{IEEEtran}
\bibliography{reference}

\begin{thebibliography}{10}
\providecommand{\url}[1]{#1}
\csname url@samestyle\endcsname
\providecommand{\newblock}{\relax}
\providecommand{\bibinfo}[2]{#2}
\providecommand{\BIBentrySTDinterwordspacing}{\spaceskip=0pt\relax}
\providecommand{\BIBentryALTinterwordstretchfactor}{4}
\providecommand{\BIBentryALTinterwordspacing}{\spaceskip=\fontdimen2\font plus
\BIBentryALTinterwordstretchfactor\fontdimen3\font minus
  \fontdimen4\font\relax}
\providecommand{\BIBforeignlanguage}[2]{{%
\expandafter\ifx\csname l@#1\endcsname\relax
\typeout{** WARNING: IEEEtran.bst: No hyphenation pattern has been}%
\typeout{** loaded for the language `#1'. Using the pattern for}%
\typeout{** the default language instead.}%
\else
\language=\csname l@#1\endcsname
\fi
#2}}
\providecommand{\BIBdecl}{\relax}
\BIBdecl

\bibitem{karpathy2014deep}
A.~Karpathy, A.~Joulin, and F.~F.~F. Li, ``Deep fragment embeddings for
  bidirectional image sentence mapping,'' in \emph{NIPS}, 2014.

\bibitem{klein2014fisher}
B.~Klein, G.~Lev, G.~Sadeh, and L.~Wolf, ``Fisher vectors derived from hybrid
  gaussian-laplacian mixture models for image annotation,'' \emph{CVPR}, 2015.

\bibitem{gong2014improving}
Y.~Gong, L.~Wang, M.~Hodosh, J.~Hockenmaier, and S.~Lazebnik, ``Improving
  image-sentence embeddings using large weakly annotated photo collections,''
  in \emph{ECCV}, 2014.

\bibitem{karpathy2015deep}
A.~Karpathy and L.~Fei-Fei, ``Deep visual-semantic alignments for generating
  image descriptions,'' in \emph{CVPR}, 2015.

\bibitem{johnson2015densecap}
J.~Johnson, A.~Karpathy, and L.~Fei-Fei, ``Densecap: Fully convolutional
  localization networks for dense captioning,'' \emph{CVPR}, 2016.

\bibitem{vinyals2015show}
O.~Vinyals, A.~Toshev, S.~Bengio, and D.~Erhan, ``Show and tell: A neural image
  caption generator,'' in \emph{CVPR}, 2015.

\bibitem{antol2015vqa}
S.~Antol, A.~Agrawal, J.~Lu, M.~Mitchell, D.~Batra, C.~Lawrence~Zitnick, and
  D.~Parikh, ``Vqa: Visual question answering,'' in \emph{ICCV}, 2015.

\bibitem{yu2015visual}
L.~Yu, E.~Park, A.~C. Berg, and T.~L. Berg, ``Visual madlibs: Fill in the blank
  image generation and question answering,'' \emph{ICCV}, 2015.

\bibitem{jabri2016revisiting}
A.~Jabri, A.~Joulin, and L.~van~der Maaten, ``Revisiting visual question
  answering baselines,'' in \emph{ECCV}, 2016.

\bibitem{kazemzadeh2014referitgame}
S.~Kazemzadeh, V.~Ordonez, M.~Matten, and T.~L. Berg, ``Referitgame: Referring
  to objects in photographs of natural scenes.'' in \emph{EMNLP}, 2014.

\bibitem{yu2016modeling}
L.~Yu, P.~Poirson, S.~Yang, A.~C. Berg, and T.~L. Berg, ``Modeling context in
  referring expressions,'' in \emph{ECCV}, 2016.

\bibitem{plummer2015flickr30k}
B.~A. Plummer, L.~Wang, C.~M. Cervantes, J.~C. Caicedo, J.~Hockenmaier, and
  S.~Lazebnik, ``Flickr30k entities: Collecting region-to-phrase
  correspondences for richer image-to-sentence models,'' in \emph{ICCV}, 2015.

\bibitem{chen2015microsoft}
X.~Chen, H.~Fang, T.-Y. Lin, R.~Vedantam, S.~Gupta, P.~Doll{\'a}r, and C.~L.
  Zitnick, ``Microsoft coco captions: Data collection and evaluation server,''
  \emph{arXiv preprint arXiv:1504.00325}, 2015.

\bibitem{young2014image}
P.~Young, A.~Lai, M.~Hodosh, and J.~Hockenmaier, ``From image descriptions to
  visual denotations: New similarity metrics for semantic inference over event
  descriptions,'' \emph{Transactions of the Association for Computational
  Linguistics}, vol.~2, pp. 67--78, 2014.

\bibitem{krishnavisualgenome}
\BIBentryALTinterwordspacing
R.~Krishna, Y.~Zhu, O.~Groth, J.~Johnson, K.~Hata, J.~Kravitz, S.~Chen,
  Y.~Kalantidis, L.-J. Li, D.~A. Shamma, M.~Bernstein, and L.~Fei-Fei, ``Visual
  genome: Connecting language and vision using crowdsourced dense image
  annotations,'' 2016. [Online]. Available:
  \url{https://arxiv.org/abs/1602.07332}
\BIBentrySTDinterwordspacing

\bibitem{mikolov2013distributed}
T.~Mikolov, I.~Sutskever, K.~Chen, G.~S. Corrado, and J.~Dean, ``Distributed
  representations of words and phrases and their compositionality,'' in
  \emph{NIPS}, 2013.

\bibitem{lin2014microsoft}
T.-Y. Lin, M.~Maire, S.~Belongie, J.~Hays, P.~Perona, D.~Ramanan,
  P.~Doll{\'a}r, and C.~L. Zitnick, ``Microsoft coco: Common objects in
  context,'' in \emph{ECCV}, 2014.

\bibitem{eisenschtat2016linking}
A.~Eisenschtat and L.~Wolf, ``Linking image and text with 2-way nets,''
  \emph{CVPR}, 2017.

\bibitem{wang2015learning}
L.~Wang, Y.~Li, and S.~Lazebnik, ``Learning deep structure-preserving
  image-text embeddings,'' \emph{CVPR}, 2016.

\bibitem{yu2016joint}
L.~Yu, H.~Tan, M.~Bansal, and T.~L. Berg, ``A joint speaker-listener-reinforcer
  model for referring expressions,'' \emph{CVPR}, 2017.

\bibitem{hardoon2004canonical}
D.~R. Hardoon, S.~Szedmak, and J.~Shawe-Taylor, ``Canonical correlation
  analysis: An overview with application to learning methods,'' \emph{Neural
  computation}, vol.~16, no.~12, pp. 2639--2664, 2004.

\bibitem{hotelling1936}
H.~Hotelling, ``Relations between two sets of variables,'' \emph{Biometrika},
  vol.~28, p. 312–377, 1936.

\bibitem{gong2014multi}
Y.~Gong, Q.~Ke, M.~Isard, and S.~Lazebnik, ``A multi-view embedding space for
  modeling internet images, tags, and their semantics,'' \emph{IJCV}, 2014.

\bibitem{hodosh2013framing}
M.~Hodosh, P.~Young, and J.~Hockenmaier, ``Framing image description as a
  ranking task: Data, models and evaluation metrics,'' \emph{Journal of
  Artificial Intelligence Research}, 2013.

\bibitem{andrew2013deep}
G.~Andrew, R.~Arora, J.~Bilmes, and K.~Livescu, ``Deep canonical correlation
  analysis,'' in \emph{ICML}, 2013.

\bibitem{mikolajczyk2015deep}
F.~Yan and K.~Mikolajczyk, ``Deep correlation for matching images and text,''
  in \emph{CVPR}, 2015.

\bibitem{ma2015finding}
Z.~Ma, Y.~Lu, and D.~Foster, ``Finding linear structure in large datasets with
  scalable canonical correlation analysis,'' \emph{ICML}, 2015.

\bibitem{ngiam2011multimodal}
J.~Ngiam, A.~Khosla, M.~Kim, J.~Nam, H.~Lee, and A.~Y. Ng, ``Multimodal deep
  learning,'' in \emph{ICML}, 2011.

\bibitem{srivastava2012multimodal}
N.~Srivastava and R.~R. Salakhutdinov, ``Multimodal learning with deep
  boltzmann machines,'' in \emph{NIPS}, 2012.

\bibitem{donahue2014long}
J.~Donahue, L.~A. Hendricks, S.~Guadarrama, M.~Rohrbach, S.~Venugopalan,
  K.~Saenko, and T.~Darrell, ``Long-term recurrent convolutional networks for
  visual recognition and description,'' \emph{arXiv:1411.4389}, 2014.

\bibitem{kiros2014multimodal}
R.~Kiros, R.~Salakhutdinov, and R.~S. Zemel, ``Multimodal neural language
  models.'' in \emph{ICML}, 2014.

\bibitem{kiros2014unifying}
R.~Kiros, R.~Salakhutdinov, and R.~Zemel, ``Unifying visual-semantic embeddings
  with multimodal neural language models,'' in \emph{arXiv preprint
  arXiv:1411.2539}, 2014.

\bibitem{mao2014deep}
J.~Mao, W.~Xu, Y.~Yang, J.~Wang, and A.~Yuille, ``Deep captioning with
  multimodal recurrent neural networks (m-rnn),'' \emph{ICLR}, 2015.

\bibitem{venugopalan2014translating}
S.~Venugopalan, H.~Xu, J.~Donahue, M.~Rohrbach, R.~Mooney, and K.~Saenko,
  ``Translating videos to natural language using deep recurrent neural
  networks,'' \emph{arXiv:1412.4729}, 2014.

\bibitem{weston2011wsabie}
J.~Weston, S.~Bengio, and N.~Usunier, ``Wsabie: Scaling up to large vocabulary
  image annotation,'' in \emph{IJCAI}, 2011.

\bibitem{frome2013devise}
A.~Frome, G.~S. Corrado, J.~Shlens, S.~Bengio, J.~Dean, T.~Mikolov
  \emph{et~al.}, ``Devise: A deep visual-semantic embedding model,'' in
  \emph{NIPS}, 2013.

\bibitem{socher2014grounded}
R.~Socher, A.~Karpathy, Q.~V. Le, C.~D. Manning, and A.~Y. Ng, ``Grounded
  compositional semantics for finding and describing images with sentences,''
  \emph{Transactions of the Association for Computational Linguistics}, vol.~2,
  pp. 207--218, 2014.

\bibitem{hu2014discriminative}
J.~Hu, J.~Lu, and Y.-P. Tan, ``Discriminative deep metric learning for face
  verification in the wild,'' in \emph{CVPR}, 2014.

\bibitem{mensink2012metric}
T.~Mensink, J.~Verbeek, F.~Perronnin, and G.~Csurka, ``Metric learning for
  large scale image classification: Generalizing to new classes at near-zero
  cost,'' in \emph{ECCV}, 2012.

\bibitem{shaw2011learning}
B.~Shaw, B.~Huang, and T.~Jebara, ``Learning a distance metric from a
  network,'' in \emph{NIPS}, 2011.

\bibitem{shaw2009structure}
B.~Shaw and T.~Jebara, ``Structure preserving embedding,'' in \emph{ICML},
  2009.

\bibitem{weinberger2005distance}
K.~Q. Weinberger, J.~Blitzer, and L.~K. Saul, ``Distance metric learning for
  large margin nearest neighbor classification,'' in \emph{NIPS}, 2005.

\bibitem{vzbontar2014computing}
J.~{\v{Z}}bontar and Y.~LeCun, ``Computing the stereo matching cost with a
  convolutional neural network,'' \emph{arXiv:1409.4326}, 2014.

\bibitem{bromley1993signature}
J.~Bromley, J.~W. Bentz, L.~Bottou, I.~Guyon, Y.~LeCun, C.~Moore,
  E.~S{\"a}ckinger, and R.~Shah, ``Signature verification using a “siamese”
  time delay neural network,'' \emph{International Journal of Pattern
  Recognition and Artificial Intelligence}, vol.~7, no.~04, pp. 669--688, 1993.

\bibitem{chopra2005learning}
S.~Chopra, R.~Hadsell, and Y.~LeCun, ``Learning a similarity metric
  discriminatively, with application to face verification,'' in \emph{CVPR},
  2005.

\bibitem{han2015matchnet}
X.~Han, T.~Leung, Y.~Jia, R.~Sukthankar, and A.~C. Berg, ``Matchnet: Unifying
  feature and metric learning for patch-based matching,'' in \emph{CVPR}, 2015.

\bibitem{hoffer2014deep}
E.~Hoffer and N.~Ailon, ``Deep metric learning using triplet network,''
  \emph{ICLR}, 2015.

\bibitem{schroff2015facenet}
F.~Schroff, D.~Kalenichenko, and J.~Philbin, ``Facenet: A unified embedding for
  face recognition and clustering,'' \emph{CVPR}, 2015.

\bibitem{wang2014learning}
J.~Wang, Y.~Song, T.~Leung, C.~Rosenberg, J.~Wang, J.~Philbin, B.~Chen, and
  Y.~Wu, ``Learning fine-grained image similarity with deep ranking,'' in
  \emph{CVPR}, 2014.

\bibitem{ba2015predicting}
J.~Ba, K.~Swersky, S.~Fidler, and R.~Salakhutdinov, ``Predicting deep zero-shot
  convolutional neural networks using textual descriptions,'' \emph{ICCV},
  2015.

\bibitem{fukui2016multimodal}
A.~Fukui, D.~H. Park, D.~Yang, A.~Rohrbach, T.~Darrell, and M.~Rohrbach,
  ``Multimodal compact bilinear pooling for visual question answering and
  visual grounding,'' \emph{arXiv preprint arXiv:1606.01847}, 2016.

\bibitem{rohrbach2015grounding}
A.~Rohrbach, M.~Rohrbach, R.~Hu, T.~Darrell, and B.~Schiele, ``Grounding of
  textual phrases in images by reconstruction,'' \emph{ECCV}, 2016.

\bibitem{zitnick2014edge}
C.~L. Zitnick and P.~Doll{\'a}r, ``Edge boxes: Locating object proposals from
  edges,'' in \emph{ECCV}, 2014.

\bibitem{ioffe2015batch}
S.~Ioffe and C.~Szegedy, ``Batch normalization: Accelerating deep network
  training by reducing internal covariate shift,'' \emph{ICML}, 2015.

\bibitem{joachims2009cutting}
T.~Joachims, T.~Finley, and C.-N.~J. Yu, ``Cutting-plane training of structural
  svms,'' \emph{Machine Learning}, vol.~77, no.~1, pp. 27--59, 2009.

\bibitem{plummer2016ijcv}
B.~Plummer, L.~Wang, C.~Cervantes, J.~Caicedo, J.~Hockenmaier, and S.~Lazebnik,
  ``Flickr30k entities: Collecting region-to-phrase correspondences for richer
  image-to-sentence models,'' \emph{IJCV}, 2016.

\bibitem{girshick2015fast}
R.~Girshick, ``Fast r-cnn,'' in \emph{ICCV}, 2015.

\bibitem{simonyan2014very}
K.~Simonyan and A.~Zisserman, ``Very deep convolutional networks for
  large-scale image recognition,'' \emph{arXiv:1409.1556}, 2014.

\bibitem{everingham2011pascal}
M.~Everingham, L.~Van~Gool, C.~Williams, J.~Winn, and A.~Zisserman, ``The
  pascal visual object classes challenge 2012,'' 2011.

\bibitem{perronnin2010improving}
F.~Perronnin, J.~Sanchez, and T.~Mensink, ``Improving the {F}isher kernel for
  large-scale image classification,'' in \emph{ECCV}, 2010.

\bibitem{wang2016structured}
M.~Wang, M.~Azab, N.~Kojima, R.~Mihalcea, and J.~Deng, ``Structured matching
  for phrase localization,'' in \emph{ECCV}, 2016.

\bibitem{kingma2014adam}
D.~Kingma and J.~Ba, ``Adam: A method for stochastic optimization,''
  \emph{arXiv preprint arXiv:1412.6980}, 2014.

\bibitem{ma2015multimodal}
L.~Ma, Z.~Lu, L.~Shang, and H.~Li, ``Multimodal convolutional neural networks
  for matching image and sentence,'' \emph{ICCV}, 2015.

\bibitem{krizhevsky2012imagenet}
A.~Krizhevsky, I.~Sutskever, and G.~E. Hinton, ``Image{N}et classification with
  deep convolutional neural networks,'' in \emph{NIPS}, 2012.

\bibitem{ba2016layer}
J.~L. Ba, J.~R. Kiros, and G.~E. Hinton, ``Layer normalization,'' \emph{arXiv
  preprint arXiv:1607.06450}, 2016.

\bibitem{vendrov2015order}
I.~Vendrov, R.~Kiros, S.~Fidler, and R.~Urtasun, ``Order-embeddings of images
  and language,'' \emph{ICLR}, 2016.

\bibitem{hochreiter1997long}
S.~Hochreiter and J.~Schmidhuber, ``Long short-term memory,'' \emph{Neural
  computation}, vol.~9, no.~8, pp. 1735--1780, 1997.

\end{thebibliography}

\end{document}